\documentclass[letterpaper, 10 pt, conference]{ieeeconf}  % Comment this line out if you need a4paper

\IEEEoverridecommandlockouts                              % This command is only needed if 
\usepackage{graphics} % for pdf, bitmapped graphics files
\usepackage{epsfig} % for postscript graphics files
\usepackage{amsmath} % assumes amsmath package installed
\usepackage{amssymb}  % assumes amsmath package installed
\usepackage{cite}
\usepackage{xcolor}

\usepackage{subcaption}
\usepackage{caption}
\usepackage[colorlinks]{hyperref}
\usepackage{booktabs}

\title{\LARGE \bf
Predicting Like A Pilot: Dataset and Method to Predict Socially-Aware Aircraft Trajectories in Non-Towered Terminal Airspace
}

\author{Jay Patrikar$^{1}$, Brady Moon$^{1}$, Jean Oh$^{1}$ and Sebastian Scherer$^{1}$% <-this % stops a space
\thanks{*This work is supported by the U.S. Department of Energy (Grant DE-EE0008463). This material is based upon work supported by the National Science Foundation Graduate Research Fellowship under Grant No. DGE1745016.}% <-this % stops a space
\thanks{$^{1}$Authors are with the Robotics Institute, Carnegie Mellon University, Pittsburgh, PA, USA.
        {\tt\small \{jpatrika, bradym, hyaejino, basti\}@andrew.cmu.edu}}%
}

\begin{document}

\maketitle
\thispagestyle{empty}
\pagestyle{empty}

%%%%%%%%%%%%%%%%%%%%%%%%%%%%%%%%%%%%%%%%%%%%%%%%%%%%%%%%%%%%%%%%%%%%%%%%%%%%%%%%
\begin{abstract}

Pilots operating aircraft in non-towered terminal airspace rely on their situational awareness and prior knowledge to predict the future trajectories of other agents. These predictions are conditioned on the past trajectories of other agents, agent-agent social interactions and environmental context such as airport location and weather. This paper provides a dataset, \textit{TrajAir}, that captures this behaviour in non-towered terminal airspace around a regional airport. We also present a baseline socially-aware trajectory prediction algorithm, \textit{TrajAirNet}, that uses the dataset to predict the trajectories of all agents. The dataset is collected for 111 days over 8 months and contains ADS-B transponder data along with the corresponding METAR weather data. The data is processed to be used as a benchmark with other publicly available social navigation datasets. To the best of the authors' knowledge, this is the first 3D social aerial navigation dataset, thus introducing social navigation for autonomous aviation. \textit{TrajAirNet} combines state-of-the-art modules in social navigation to provide predictions in a static environment with a dynamic context. Both the \textit{TrajAir} dataset and \textit{TrajAirNet} prediction algorithm are open-source.\\ \href{https://theairlab.org/trajair/}{[Dataset]}\footnote{\href{https://theairlab.org/trajair/}{Dataset: https://theairlab.org/trajair/}} \href{https://github.com/castacks/trajairnet}{[Code]}\footnote{ \href{https://github.com/castacks/trajairnet}{Codebase: https://github.com/castacks/trajairnet}}  
\href{https://youtu.be/elAQXrxB2gw}{[Video]}\footnote{ \href{https://youtu.be/elAQXrxB2gw}{Video: https://youtu.be/elAQXrxB2gw}}  

\end{abstract}

%%%%%%%%%%%%%%%%%%%%%%%%%%%%%%%%%%%%%%%%%%%%%%%%%%%%%%%%%%%%%%%%%%%%%%%%%%%%%%%%
\section{INTRODUCTION}

General Aviation (GA) comprises all civil flights except scheduled passenger airline services. More than 90\% of the roughly 220,000 civil aircraft registered in the United States (US) are GA aircraft \cite{federal-aviation-administration-2018}. In contrast with airline service aircraft, which operate with two pilots in a structured higher-altitude operational envelope, GA aircraft are often individually piloted in a more unstructured lower-altitude environment. This makes the pilotage challenging in the best of circumstances. According to the Federal Aviation Administration (FAA), for every commercial airline accident in 2015, there were approximately 36 accidents in GA, with 77 \% of non-fatal accidents in terminal airspaces \cite{national-transportation-safety-board-2015}. This low altitude environment is also where a bulk of the next generation of Unmanned Aerial Vehicles (UAVs) are expected to operate \cite{prevot2016uas}. These UAVs are expected to seamlessly interact with other UAVs and crewed air traffic operating in this shared airspace. Nowhere is this interaction more pronounced than in low-altitude terminal airspace around airports.

% \begin{figure}[th]
%     \centering
%     \begin{subfigure}[b]{0.45\textwidth}
%          \centering
%         \includegraphics[width=\textwidth,trim={0cm 0cm 0cm 0cm},clip]{figures/Fig1av8.png}
%             \caption{}
%          \label{fig:result_b}
%      \end{subfigure}
%      \begin{subfigure}[b]{0.45\textwidth}
%          \centering
%         \includegraphics[trim={0cm 0cm 0cm 0cm},clip,width=\textwidth]{figures/Fig1bv3.pdf}
%     \caption{}
%          \label{fig:result_b}
%      \end{subfigure}
%         \caption{For a given input trajectory and environmental context, \textit{TrajAirNet}  returns a distribution of future trajectories. }
%     \label{fig:overview}
% \end{figure}

\begin{figure}[th]
    \centering
         \centering
        \includegraphics[width=.5\textwidth,trim={0cm 0cm -.8cm -.5cm},clip]{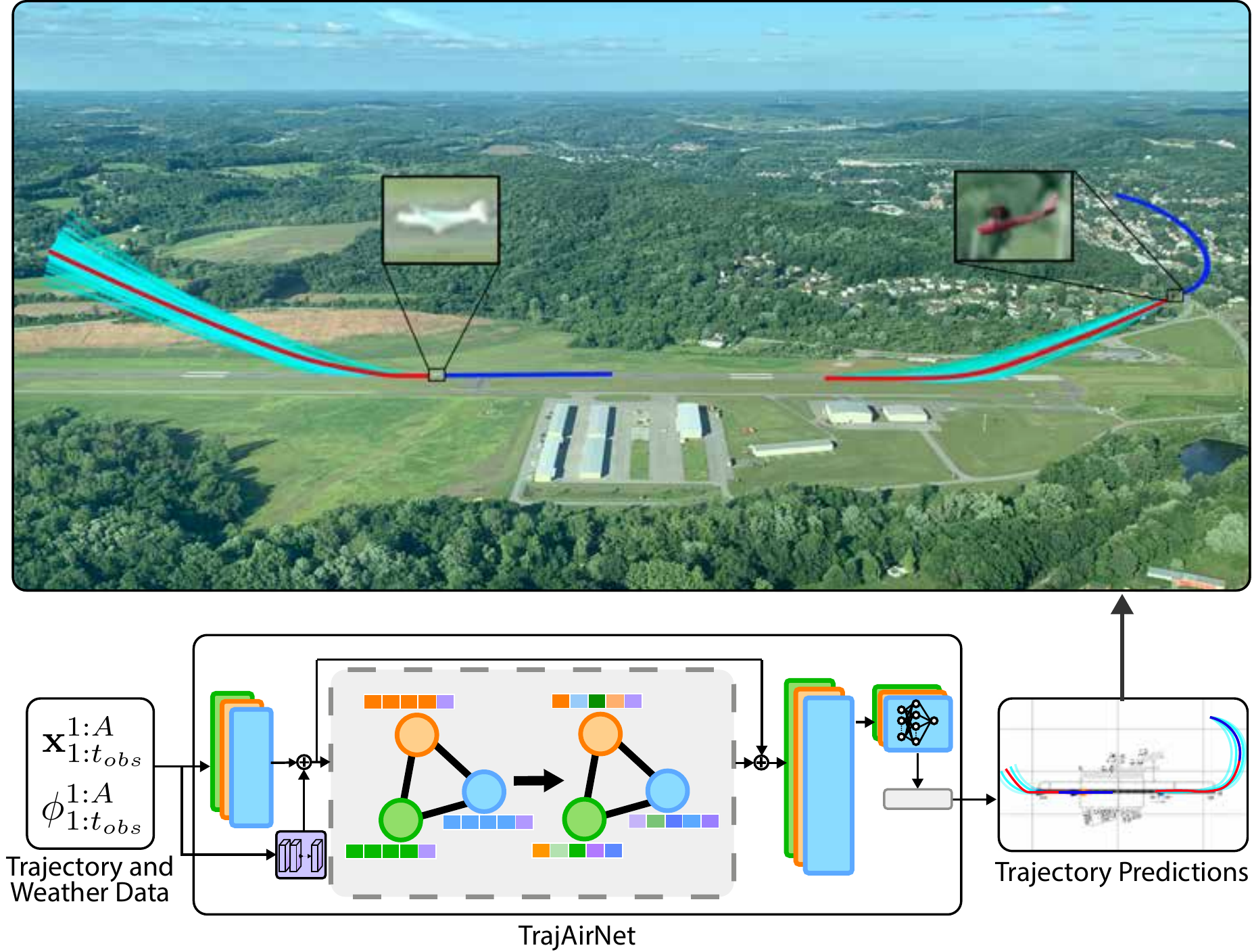}
        \caption{ \textit{TrajAir} dataset includes recorded ADS-B trajectories of aircraft interacting in a non-towered general aviation airport and the weather context from METAR strings. \textit{TrajAirNet} predicts multi-future samples (cyan) of trajectories for all agents by conditioning each on the history (blue) of all the agents and the weather context.}
        \vspace{-3.5mm}
    \label{fig:overview}
\end{figure}

All flights typically begin and end in airspace surrounding airports known as terminal airspace. Low altitudes, multi-agent close-proximity interactions, dynamically changing conditions, and rapid decision making are hallmarks of this type of airspace. As compared to en-route airspace, where agents are typically well-separated, agents in terminal airspace are at a higher collision risk. Out of the nearly 20,000 active airports \cite{johnson2017estimating} in the US, only around 4\% are \textit{towered}, meaning that a control tower is present as a centralized authority ensuring separation. This indicates that most GA airports are \textit{non-towered}, implying that the pilots must directly communicate with other pilots and take decentralized actions. 
Pilots operating in non-towered airspace are solely responsible for guiding aircraft to safety. 
\begin{figure*}[!hbt]
    \centering
        \includegraphics[width=\textwidth]{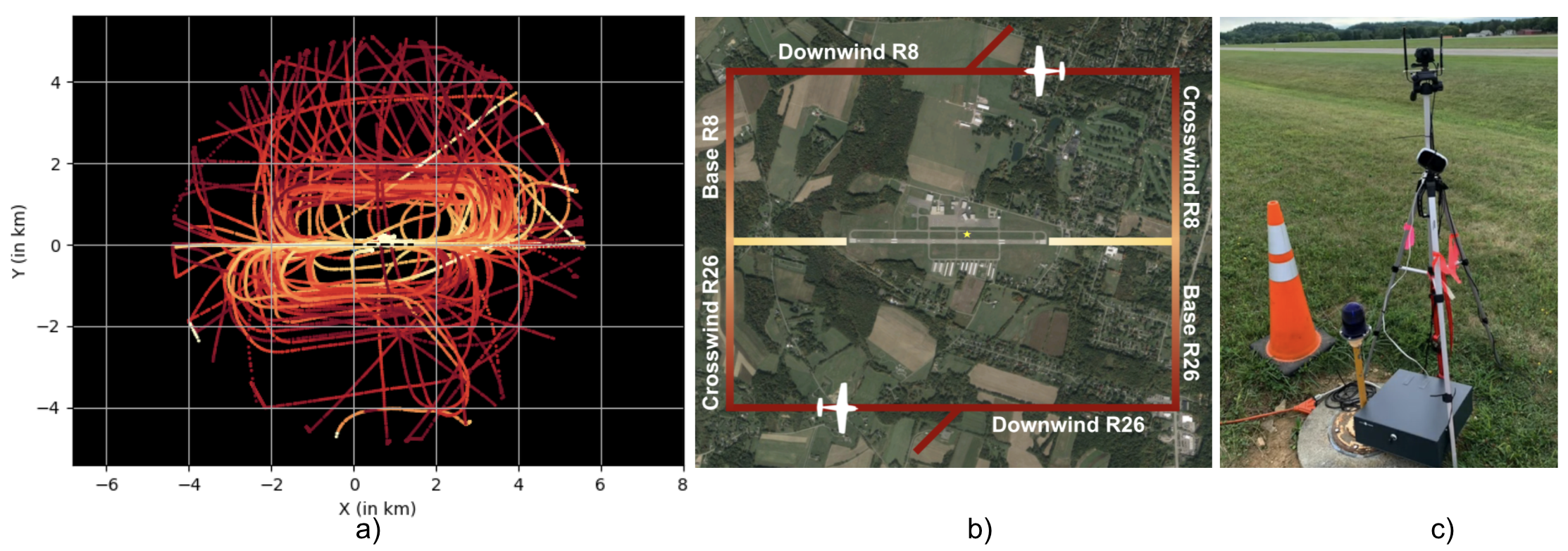}
    \caption{Figure shows the dataset and its collection setup at the Pittsburgh-Butler Regional Airport (KBTP)---a non-towered GA airport that serves as a primary location for the dataset. Lighter color indicates lower altitude. (a) Shown is a snippet of the processed dataset with aircraft trajectories showing clear lobes for traffic patterns for both runways. (b) The left traffic pattern and nomenclature for the runways at the airport. (c) Picture of the data-collection setup.}
    \label{kbtp}
\end{figure*}

In the context of social navigation, where only implicit rules are assumed~\cite{mavrogiannis2021core}, GA offers a unique case where all aircraft operating within terminal airspace are expected to self-organise to follow guidelines established by the FAA, which act like rules-of-the-road for aircraft. These guidelines ensure separation and smooth flow by standardising a rectangular traffic pattern around the runways \cite{federal-aviation-administration-2018B}. More details on these guidelines are provided in future sections. While not enforced, pilots are expected to adhere to the traffic pattern, but deviations arise due to the lack of specificity, pilot's experience level, type of aircraft, weather, and position of other agents. This design flexibility improves efficiency and ensures safety while accommodating the needs of various agents using the same infrastructure. Only the traffic pattern's general shape, direction, and altitude are established. This leaves room for each pilot to make their own decisions to maintain separation while respecting personal minimums and aircraft capabilities. 

Airports can have multiple agents using the same airspace where all agents are expected to follow socially-compliant trajectories while loosely following the traffic pattern guidelines. Using a diverse set of inputs such as radio, transponder data, weather, and vision, each pilot constructs their airspace situational awareness  to decide an entry or exit from the pattern. To avoid collisions, therefore, it is critical for each pilot to predict the future trajectory predictions of other agents in that airspace.  
%Pilots use diverse set of inputs such as radio, transponder data, weather, and vision to gain a situational awareness of the airspace. Thus, an entry and exit from the pattern is based on the pilot's assessment of the current situation and often pilot's modify their trajectories based on the future trajectory predictions of other agents in that airspace. Airports can have multiple agents using the same airspace and each is expected to follow socially-compliant trajectories while loosely following the traffic pattern guidelines. Thus
In this work, we address the problem of learning how to predict the actions of other agents (aircraft) in a static environment with the dynamic context needed to navigate an aircraft in non-towered terminal airspace safely.

% Understanding this airspace provides two benefits. One obvious goal is to help human pilots augment their situational awareness to improve their decision making. Tools can be designed that can provide pilots a heightened situational awareness by providing a prediction on each of the agents sharing the airspace. A second, more long-term goal, is to help unmanned aircraft achieve a better understanding of this airspace so as to seamlessly integrate them into the existing National Airspace (NAS).

 This work presents a novel dataset, \textit{TrajAir}, that provides recorded trajectories of multiple aircraft operating around a standard non-towered airport while also providing the weather conditions during these operations. While several datasets exist for ground navigation or pedestrian trajectories, few datasets are available for aerial navigation that imposes unique challenges including the need to handle regulations and weather conditions. 
 
 This work also provides a baseline trajectory prediction method, \textit{TrajAirNet}, to model agent-agent, agent-environment, and agent-context interactions. The future trajectories of agents around an airport are a function of the history of all the participating heterogeneous agents, their respective goals, the static airport environment, and the dynamic weather context. The proposed method uses the dataset to effectively capture all aspects of this behaviour to predict aircraft trajectories. 
%  The proposed dataset and method provides access to a novel arena with heterogeneous agents interacting in a static environment with a dynamic context.

The major contributions of this work are as follows:

\begin{enumerate}
    \item We provide the first publicly-available large-scale processed trajectory and weather data with multiple aircraft socially interacting in a non-towered GA airport.  
    \item We provide an open-source end-to-end attention-based baseline method to predict socially-aware multi-future trajectories in a static environment with a dynamic context. 
\end{enumerate}

The paper is organised as follows: Section \ref{sec:related_works} provides details on prior work in datasets and methods in the aircraft trajectory prediction domain, along with a brief background on the pedestrian and autonomous vehicle domains. Section \ref{sec:trajair} provides details on the \textit{TrajAir} dataset, while Section \ref{sec:trajairnet} discusses the \textit{TrajAirNet} method. Section \ref{sec:eval} discusses the metrics and provides qualitative and quantitative results. Sections \ref{sec:future} and \ref{sec:conclusion} discuss the future work and conclusions, respectively.

\section{Related Work}\label{sec:related_works}
\subsection{Aircraft Trajectory Prediction}
Previous work on aircraft trajectory prediction can be split into macro-level predictions in high-altitude en-route airspace and micro-level predictions for terminal airspace. En-route long-range predictions are often a function of weather conditions \cite{ayhan2016aircraft,pang2019recurrent,franco2018probabilistic,pang2020conditional}. For terminal airspace, previous work has focused on larger airports with commercial aviation (CA) traffic. CA aircraft usually follow strict approach procedures for entering and exiting terminal airspace and are often guided by air traffic control, making the trajectory prediction problem akin to learning the statistical variation in following these procedures \cite{zeng2020deep,barratt2018learning,alligier2018learning}. Non-towered terminal airspace, on the other hand, while still following FAA guidelines, has a much higher complexity and often expects air traffic to follow socially compliant behaviour. The lack of a centralised authority gives the pilots more freedom with the choice of actions to achieve a particular goal. To the best of the authors' knowledge, no publicly available dataset or open-source trajectory prediction methods exist for non-towered GA traffic, which forms the bulk of the aviation infrastructure.       
\subsection{Trajectory Prediction Datasets}
 The majority of the research in socially-aware human-robot interaction has focused on pedestrians and autonomous vehicles (AVs) \cite{deo2020trajectory,mangalam2020not,marchetti2020mantra,salzmann2020trajectron++,yuan2021agentformer}. The arena is often crowded spaces like shopping malls, college campuses, or city streets. Pedestrian datasets like UCY \cite{lerner2007crowds}, ETH \cite{pellegrini2010improving}, and the Stanford Drone Dataset \cite{robicquet2016learning}, among others \cite{rudenko2020human}, have long been the dominant datasets for evaluating pedestrian trajectory prediction tasks. Within the AV domain, Argoverse \cite{chang2019argoverse}, KITTI \cite{geiger2012we} and nuScenes \cite{caesar2020nuscenes} have been more popular.

 With \textit{TrajAir}, we introduce the new 3D domain of general aviation within the paradigm of social navigation. The dataset differs from previously published datasets due to the spatial dependence of the agent trajectories within a static environment. \textit{TrajAir} trajectories are conditioned not only on the relative location of agents but also on their absolute locations. In addition, trajectories are explicitly goal-directed. This enables benchmarking models that use goal- or intention-driven predictions. Due to the FAA guidelines, the trajectories loosely follow the same semantic structure opening up the field of structured predictions in social navigation. Another point of difference is the presence of a global context in the form of weather data directly affects the trajectories and goals of all agents. This allows benchmarking algorithms that can use contextual clues to aid social behaviour prediction.

\subsection{Trajectory Prediction Algorithms}
% Predicting socially-aware trajectories of pedestrians and AVs is a long standing problem in robotics \cite{yuan2021agentformer,deo2020trajectory,salzmann2020trajectron++,mangalam2020goals,mangalam2020not,marchetti2020mantra}.
Social trajectory prediction algorithms typically use three separable modules to generate trajectory predictions. A sequential module like recurrent units \cite{lee2017desire} or convolutions \cite{nikhil2018convolutional} is first used to encode the observed trajectories of each agent to generate a vector in latent space. A social module then uses a method like pooling \cite{alahi2016social} or attention \cite{mohamed2020social} to encode the social context. A generative module then decodes the socially-aware representation into an output that is either the relative coordinates \cite{lee2017desire}, a distribution on the relative coordinates \cite{zhao2020noticing} or accelerations \cite{rhinehart2019precog}. Due to these algorithms' the domain specific nature they do not generalize well to GA aircraft. The use of relative or ego-centric coordinate systems, suitable for pedestrians and AV domains, renders these algorithms unsuitable for domains like GA, where the decisions are a function of the absolute spatial coordinates. While algorithms like Trajectron++\cite{salzmann2020trajectron++} have been proposed to support domain independence with support for dynamics, its extension to 3D space, airport maps and double-integrator dynamics is non-trivial. 
The proposed baseline method, \textit{TrajAirNet}, which uses absolute coordinates and global contexts, builds on similar structuring to provide a trajectory prediction algorithm that combines the state-of-the-art in each of the aforementioned modules.       

\section{\textit{TrajAir} Dataset}\label{sec:trajair}

% \begin{figure*}
%      \centering
%      \begin{subfigure}[b]{0.37\textwidth}
%          \centering
%     \includegraphics[width=\textwidth]{figures/data.png}
%     \caption{}
%          \label{fig:y equals x}
%      \end{subfigure}
%      \begin{subfigure}[b]{0.31\textwidth}
%          \centering
%         \includegraphics[width=\textwidth]{figures/pattern.png}
%             \caption{}
%          \label{fig:three sin x}
%      \end{subfigure}
%      \begin{subfigure}[b]{0.2\textwidth}
%          \centering
%     \includegraphics[width=\textwidth]{figures/setup.png}
%         \caption{}
%          \label{fig:five over x}
%      \end{subfigure}
% \caption{\todoBrady{Fix} Figure shows the dataset and its collection setup at the Pittsburgh-Butler Regional Airport; a non-towered GA airport which serves as a primary location for the dataset. (a) Shown is a snippet of the processed dataset with aircraft trajectories showing clear lobes for traffic patterns for both runways. Lighter color indicates lower altitude. (b) The star-marker shows the approximate location of the data collection setup at the airport. (c) File-photo of the data-collection setup. }
%     \label{fig:my_label}        \label{fig:three graphs}
% \end{figure*}

The \textit{TrajAir} dataset is collected at the Pittsburgh-Butler Regional Airport (ICAO:KBTP), a single runway GA airport, 10 miles North of the city of Pittsburgh, Pennsylvania. Additional information about KBTP is available online\footnote{\url{http://www.airnav.com/airport/kbtp}}. Aircraft entering and leaving non-towered airspace need to follow guidelines established by the FAA to ensure the safety and efficiency of all participating agents. KBTP has Left Traffic patterns for both runways, meaning the patterns are rectangular-shaped with left-handed turns relative to the direction of landing or take-off. Figure~\ref{kbtp}b shows the traffic pattern for Runway 8 and 26 around KBTP with the corresponding direction of traffic flow. Aircraft usually take-off or land into the wind; hence the nomenclature follows this sequence. When an aircraft takes off, it is on an upwind leg. A left turn puts it on a crosswind leg, followed by turns into downwind leg and base leg. The final left turn puts the aircraft on the final leg for a touch-down. FAA also establishes that an entry into the pattern should be at a 45-degree angle to the downwind leg.

\begin{figure*}[t]
    \centering
        \includegraphics[width=0.9\textwidth,trim={0cm 0cm 0cm -.5cm},clip]{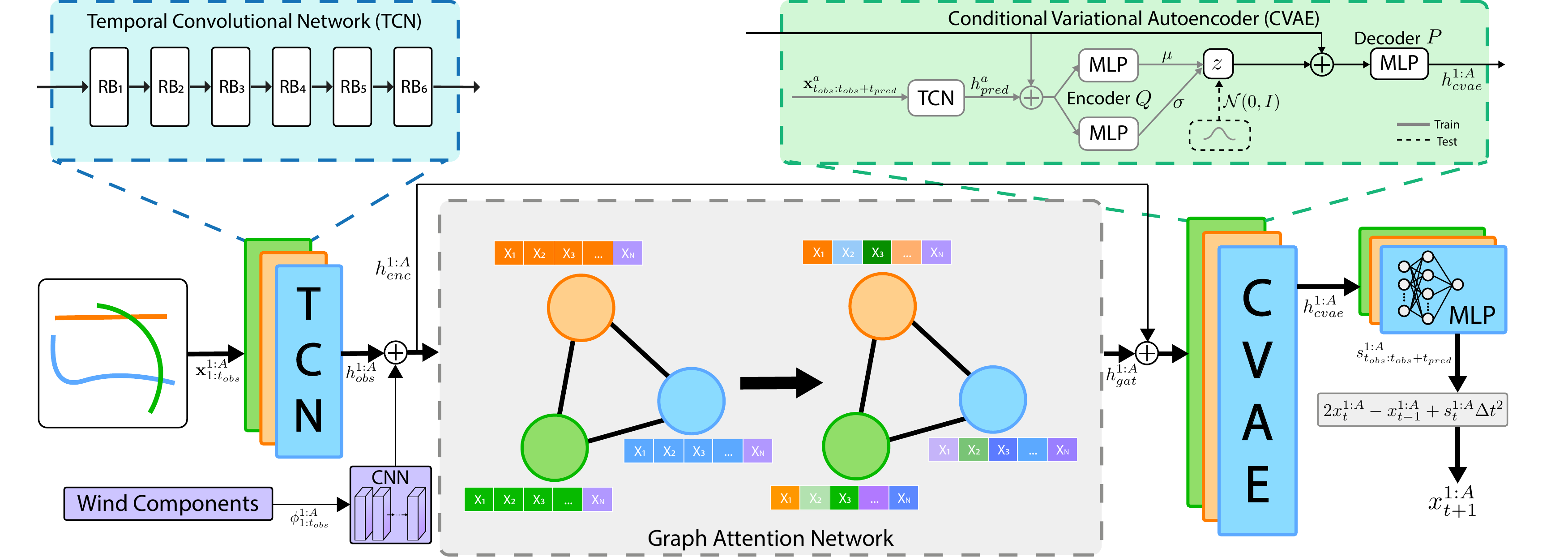}

    \caption{ Our proposed \textit{TrajAirNet} baseline model for aircraft trajectory prediction in static environment with a dynamic context. The model uses Temporal Convolutional Networks (TCNs) to encode the 3-D trajectory. The dynamic weather context (wind vectors) are encoded using Convolutional Neural Networks (CNNs), which are appended to the encoded trajectory. To encode the social context, we use a Graph Attention Network (GAT) that uses attention to combine data from different agents. Finally, we use Conditional Variational Autoencoders (CVAEs) to produce multi-future acceleration commands, which are then used in a forward Verlet integration to produce future aircraft trajectories for all agents.  }
    \label{fig:TrajAirNetModel}
\end{figure*}       

\subsection{Dataset Overview}
The trajectory data is recorded using an on-site setup (see Figure \ref{kbtp}c). Data is provided starting on 18 Sept 2020 and continues till 23 Apr 2021. It includes a total of 111 days of data discounting downtime, repairs and bad weather days with no traffic. Data is collected from 01:00 AM local time to 11:00 PM local time. The dataset can be accessed at \href{https://theairlab.org/trajair/}{https://theairlab.org/trajair/}. More information about the dataset, including the file structure and dataloaders, is also provided.  

\subsection{Trajectory Data}
The dataset uses an Automatic Dependent Surveillance-Broadcast (ADS-B) receiver \cite{dewsbury-2019} placed within the airport premises to capture the trajectory data. The ADS-B In receiver receives data directly broadcasted by other aircraft with ADS-B Out. For aircraft that do not have an ADS-B Out, the Traffic Information Service-Broadcast (TIS-B) takes the position and altitude of aircraft using radar and converts that information into a format that’s compatible with ADS-B. It then broadcasts the information to our receiver. The receiver uses both the 1090 MHz and 978 MHz frequencies to listen to these broadcasts. The ADS-B uses satellite navigation to produce accurate location and timestamp for the targets, which is recorded on-site using our custom setup.

\subsection{Weather Data}
The weather data is obtained post-hoc using the METeorological Aerodrome Reports (METAR) strings generated by the Automated Weather Observing System (AWOS) system at KBTP. We use the Iowa State METAR repository \cite{herzmann2004iowa} to gather all the weather data during the trajectory collection time frame. The raw METAR string is then appended to the raw trajectory data by matching the closest UTC timestamps.

\subsection{Data Processing}
The data obtained from the ADS-B receiver and the METAR strings is processed to make it suitable for training networks. The following steps are performed:
\begin{itemize}
    \item Removal of data points that have corrupt or no location fields.
    \item Removal of duplicate data points with the same aircraft identifier and location fields.
    \item Removal of data points where the altitude is more than 6000 feet MSL, and distance is more than 5 km from one end of the runway.
    \item Transforming the data to a local Cartesian coordinate frame in SI units. The origin is at the end of the runway, with the horizontal x-axis pointing along the runway.
    \item Processing raw METAR strings to get wind velocity and direction along and across the runway in the local Cartesian frame in SI units.
    \item Interpolating trajectory data every second for all agents.
    \item Segmenting the data into scenes with at least one active aircraft in the airport vicinity.   
    
\end{itemize}

The raw data and processed data are provided as part of the data release. A snippet of the processed data is shown in Fig \ref{kbtp}a. Lighter color indicates lower altitude. Both left-hand patterns are visible as distinct lobes on both sides of the runway. The variability in following the pattern is also evident, as are the trajectories before entering and after leaving the pattern.    
\section{\textit{TrajAirNet} Prediction Network}\label{sec:trajairnet}

We propose \textit{TrajAirNet}, an end-to-end network to provide aircraft trajectory predictions within the terminal airspace. The proposed baseline method provides us with the ability to do the following: 1) condition the prediction on ego agent's absolute position history in the static environment, 2) condition the prediction on the absolute position of other agents within the terminal airspace weighted according to their relative importance to the ego-agent, 3) condition the prediction on the dynamic environmental context like wind, 4) predict multi-modal multi-future trajectories of all the agents, and 5) operate invariant to the number of agents in the scene without using zero-padding. To achieve this, the method uses multiple components in an end-to-end fashion using a combined loss function. Section \ref{sec:problem} defines the nomenclature and provides a formal mathematical definition of the problem. Section \ref{sec:model} provides details on each model component, and Section \ref{sec:implement} provides details on the implementation and the loss function. 
\subsection{Problem Formulation}\label{sec:problem}
We consider the problem of predicting the multi-modal distribution of future trajectories of all the agents given the past history and environmental context. Let $\mathbf{x}^{a}_{t} = (x_t^a,y_t^a,z_t^a)$ and $\phi_t^a$ denote the position and context of the $a^{th}$ agent at time $t$, respectively. While the context may include terrain, maps, weather and wind conditions; for the purposes of this work, we only focus on wind as the context. Let $\mathbf{x}^{1:A}_{t_1:t_2}$ and $\phi^{1:A}_{t_1:t_2}$ denote the trajectories and context over a $\{t_1, \dots , t_2\}$ time-horizon for all $\{1, \dots, A\}$ agents in that scene. We then define our trajectory prediction problem as finding distribution of future trajectories $\mathbf{\hat{x}}^{1:A}_{t_{obs}:t_{pred}+t_{obs}}$ conditioned on the past trajectories $\mathbf{{x}}^{1:A}_{1:t_{obs}}$ and context $\phi^{1:A}_{t:t_{obs}}$, where, $t_{obs}$ is the observation time window, and $t_{pred}$ is the prediction time horizon. Mathematically,     

\begin{equation}
    \mathbf{\hat{x}}^{1:A}_{t_{obs}:t_{pred}+t_{obs}} \sim p( \mathbf{\hat{x}}^{1:A}_{t_{obs}:t_{pred}+t_{obs}} \mid \mathbf{{x}}^{1:A}_{1:t_{obs}}, \phi^{1:A}_{t:t_{obs}})
\end{equation}   
\subsection{Model Details}\label{sec:model}
We propose a baseline trajectory prediction algorithm, \textit{TrajAirNet}, that takes as input the past trajectories of all the agents, along with the weather context, to predict multi-future trajectories of all agents. The network combines elements from Temporal Convolutional Networks (TCNs), Graph Attention Network (GATs), and Conditional Variational Autoencoder (CVAEs). Each component addresses various needs of the underlying problem. The entire architecture is shown in Figure \ref{fig:TrajAirNetModel}.  

The TCN layers are used to encode the spatio-temporal trajectory into a latent vector without losing the causal relations in the underlying data. This latent representation can be used for other downstream tasks like trajectory prediction. While LSTMs have been a popular choice to encode the trajectory space, our choice of TCNs is largely because they have been shown to perform better or at least as good as LSTMs \cite{bai2018empirical}.

The GAT layer is used to encode the influence of other agents on the predicted trajectory of a particular agent. While max-pooling has been a traditional choice to incorporate the effect of other agents, the lack of interpretability in the latent encoded output makes max-pooling a rather black-box choice \cite{zhao2020noticing}. Using an attention mechanism as in GATs, on the other hand, has become a popular choice because attention can be directed selectively at particular agents in a scalable manner in terms of the number of agents~\cite{vemula2018}. This becomes especially important in the shared-airspace domain, where the relative importance of an agent is not only based on their proximity to the ego agent but also on their absolute location in space. Another advantage of using GATs is that it makes the algorithm permutation-invariant as well as invariant to the number of agents without a need to pad zeros. 

The CVAE layer serves as the backbone of the multi-future trajectory prediction. There is inherent stochasticity in the data, and CVAEs are well suited to capture this distribution. They have been used successfully in both pedestrian and AV trajectory prediction networks \cite{salzmann2020trajectron++}. The CVAEs encode the underlying distribution in its latent space, which can then be sampled at run-time to provide samples in the trajectory space. In order to incorporate the dynamics of the vehicle, we decided to a generalized dynamics formulations in the form of a forward Verlet integration \cite{rhinehart2019precog} that provides a constant velocity model for a zero output. Instead of predicting the positions directly, we predict the Verlet acceleration and then calculate the absolute positions of each agent. 

\subsection{Implementation Details}\label{sec:implement}
\textit{TrajAirNet} codebase is available open-source at \href{https://github.com/castacks/trajairnet}{https://github.com/castacks/trajairnet}. A modified dataloader breaks the scene into sequences of length $t_{obs}+t_{pred}$ with a certain minimum number of agents constant across each sequence. The number of agents can change from sequence to sequence. For each agent in a given scene, the raw trajectory in absolute coordinates is encoded using the same TCN layers. 
\begin{equation}
    h_{obs}^a = TCN_{obs} (\mathbf{{x}}^{a}_{1:t_{obs}}) ~ \forall~ a\in \{1,\dots,A\} 
\end{equation}
To include the environmental context, wind velocities along and across the runway in our case; the raw context is encoded using a standard CNN layer and the output is concatenated to the TCN encoded trajectories. The choice of CNN over MLP was motivated to enable spatial contexts in later revisions of the algorithm.  
\begin{equation}
    h_{enc}^a = h_{obs}^a\oplus CNN(\phi^{a}_{1:t_{obs}}) ~ \forall~ a\in \{1,\dots,A\} 
\end{equation}
The concatenated encoded trajectories and encoded context for all agents are then stacked together as the input to the GAT layers. Each agent acts as a node in the GAT graph structure. We use a standard GAT structure with multi-head attention \cite{velivckovic2017graph}.  
\begin{equation}
    h_{gat}^{1:A} = GAT(h_{enc}^{1:A})  
\end{equation}
The output for each agent from the GAT is then segregated and concatenated with the full encoded output $h_{enc}^a$ for each agent $a$. This concatenated vector then serves as a conditional input to the CVAE layers. The CVAE is characterized by an encoder $Q(\cdot)$ and a  decoder $P(\cdot)$. For all $a \in \{1,\dots,A\} $,
\begin{eqnarray}
h_{pred}^a &=& TCN_{pred} (\mathbf{x}^{a}_{t_{obs}:t_{obs}+t_{pred}}) \nonumber \\ 
z &\sim& Q(z \mid h_{pred}^a, h_{enc}^a \oplus h_{gat}^a), ~~ \text{for training} \nonumber \\
z &\sim& \mathcal{N}(0,I),  ~~\text{for testing} \\
h_{cvae}^a &\sim& P(h_{cvae}^a \mid z, h_{enc}^a \oplus h_{gat}^a) \nonumber
\end{eqnarray}
Finally, the sampled CVAE output is passed through a MLP layer to get the correct dimension for the acceleration output. 
\begin{equation}
    s^a_{t_{obs}:t_{obs}+t_{pred}} = MLP(h_{cvae}^a)
\end{equation}
The acceleration output is then converted to absolute positions for all $t \in \{t_{obs},\dots, t_{obs}+t_{pred}\}$, using, 
\begin{equation}
x^{1:A}_{t+1} = 2x^{1:A}_{t} -   x^{1:A}_{t-1} + s_{t}^{1:A}\Delta t^2  
\end{equation}

The entire pipeline uses a combination of loss functions. 

\begin{equation}
    \mathcal{L}_{total} = \mathcal{L}_{traj} + \mathcal{L}_{cvae}
\end{equation}
The $\mathcal{L}_{traj}$ measures how close the predicted trajectory is to the ground-truth trajectory using a mean squared error (MSE) loss.
\begin{equation}
    \mathcal{L}_{traj} = MSE(\mathbf{x}^{a}_{t_{obs}:t_{obs}+t_{pred}},\mathbf{\hat{x}}^{a}_{t_{obs}:t_{obs}+t_{pred}})
\end{equation}
The $\mathcal{L}_{cvae}$ measures the KL-Divergence between the sampling distribution of the latent variable, the easiest choice being $\mathcal{N}(0,I)$, to the distribution of latent variable that we learn during training.
\begin{equation}
    \mathcal{L}_{cvae} = D_{kl}(Q(z \mid h_{pred}^a, h_{enc}^a \oplus h_{gat}^a)\mid\mid \mathcal{N}(0,I))
\end{equation}

For training, we use the Adam optimiser with a learning rate of $1e-4$.

\section{Evaluations and Discussion}\label{sec:eval}
\begin{figure*}[h!]
     \centering
     \begin{subfigure}[b]{0.24\textwidth}
         \centering
    \includegraphics[width=\textwidth,trim={1cm 0cm 2cm .5cm},clip]{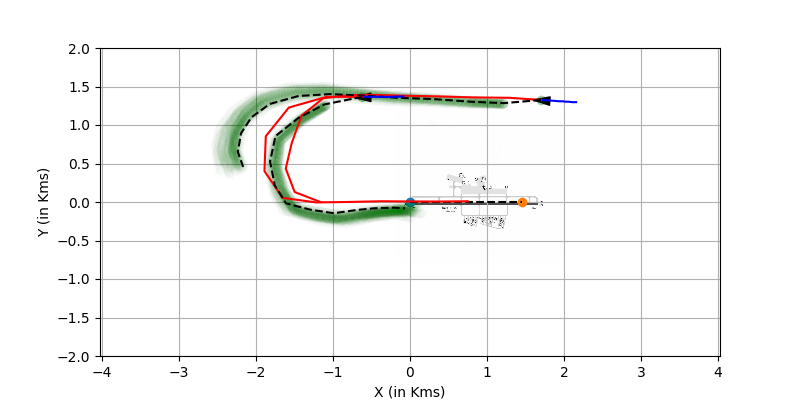}
    \caption{}
         \label{fig:result_a}
     \end{subfigure}
     \begin{subfigure}[b]{0.24\textwidth}
         \centering
        \includegraphics[width=\textwidth,trim={1cm 0cm 2cm 1cm},clip]{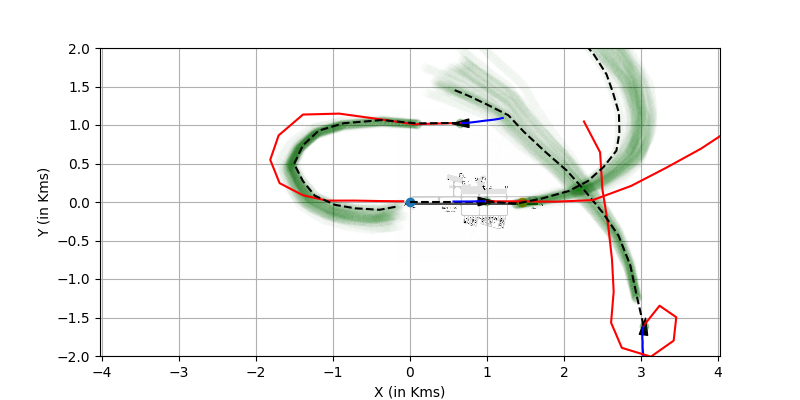}
            \caption{}
         \label{fig:result_b}
     \end{subfigure}
     \begin{subfigure}[b]{0.24\textwidth}
         \centering
    \includegraphics[width=\textwidth,trim={1cm 0cm 2cm .5cm},clip]{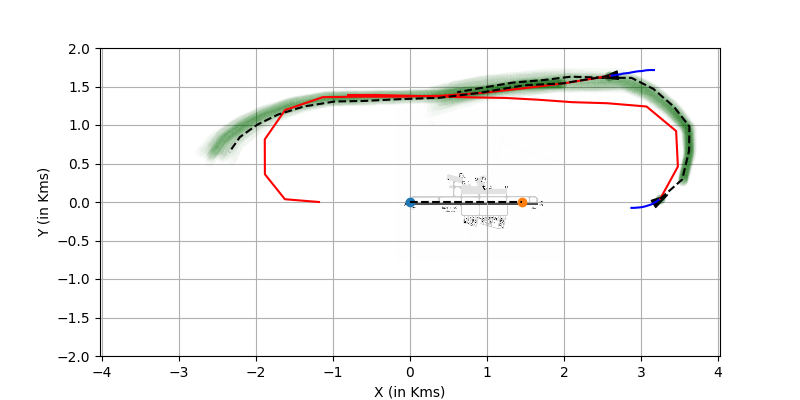}
        \caption{}
         \label{fig:result_c}
     \end{subfigure}
     \begin{subfigure}[b]{0.24\textwidth}
         \centering
    \includegraphics[width=\textwidth,trim={1cm 0cm 2cm 1cm},clip]{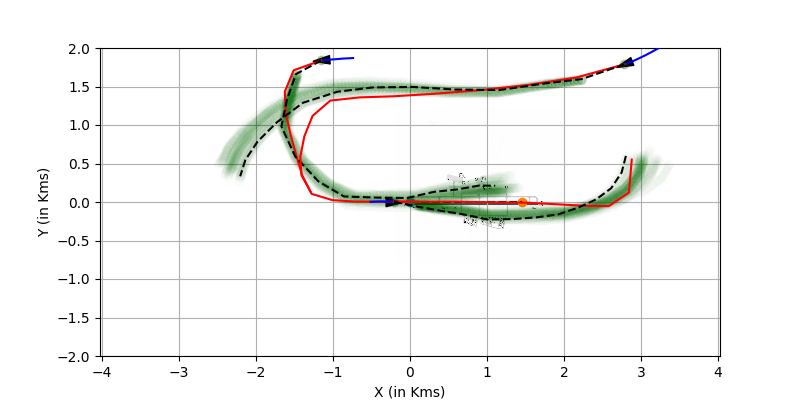}
        \caption{}
         \label{fig:result_d}
     \end{subfigure}
\caption{Figure shows the qualitative results for the \textit{TrajAirNet} framework. Four independent scenarios are chosen to showcase the performance. Blue is the observation trajectory ($t_{obs}$ = 11 sec). Green are the sampled trajectories from the \textit{TrajAirNet} output. Black shows the output closest to the ground truth ($t_{pred}$ = 120 sec), which is shown in Red. Also shown is the airport diagram to scale.}
    \label{fig:four_results}        
\end{figure*}
% \begin{figure*}
%     \centering
%     \includegraphics[width=\textwidth]{figures/res.png}
% \caption{Figure shows the qualitative results for the \textit{TrajAirNet} framework. Four independent scenarios are chosen to showcase the performance. Blue is the observation trajectory ($t_{obs}$ = 11 sec). Green are the sampled trajectories from the \textit{TrajAirNet} output. Black shows the output trajectory closest to the ground truth ($t_{pred}$ = 120 sec), which is shown in Red. Also shown is the airport diagram to scale.}       
% \label{fig:four_results}        

% \end{figure*}
\begin{table}[!ht]
    \centering
    \begin{tabular}{|l|c|c|c|c|}
    \hline
    \textbf{Algorithm}  & \textbf{7Days-1} & \textbf{7Days-2} & \textbf{7Days-3} & \textbf{7Days-4} \\
        %  \hline 
        %  & ADE/FDE & ADE/FDE & ADE/FDE & ADE/FDE \\
    \hline 
    Const. Vel.\cite{salzmann2020trajectron++} & 1.79/4.08 & 1.90/4.31 & 1.92/4.30 & 1.82/4.16  \\
    
    Nearest Neigh. & 3.13/2.70 & 1.92/1.99 & 3.41/2.69 & 2.59/2.58   \\
    
    STG-CNN\cite{mohamed2020social} & 1.19/2.35 & 1.36/2.70 & 1.33/2.67 & 1.17/2.29   \\
    TransformerTF\cite{giuliari2021transformer} & 1.58/3.85 & 1.69/4.10  &  1.97/4.36 &  1.79/4.19  \\
    \hline
    TrajAirNet (ours) & \textbf{0.73}/\textbf{1.42} & \textbf{0.81}/\textbf{1.63}&\textbf{0.86}/\textbf{1.72} & \textbf{0.71}/\textbf{1.41} \\  
    \hline
    \end{tabular}
    \caption{Table shows the quantitative results ADE/FDE (in Km) for \textit{TrajAirNet} baseline along with comparative results.}
    \label{tab:results}
\end{table}
Evaluations for the proposed network are carried out on a 28 day subset of data from the \textit{TrajAir} dataset that contains 4-sets of 7 consecutive days of data in different months. Results from our network along with comparative methods are shown for all four sets. In order to focus on long-horizon predictions, we use $t_{obs} = 11$ sec and $t_{pred} = 120 $ sec. The choice of the observation and prediction horizons are chiefly motivated to match the scale of decision horizons in GA.  
% $t_{pred}$ is then sub-sampled every 10 sec to generate a 12 step trajectory vector. A full list of network parameters is provided in the supplement. 
% \subsection{Other baselines}
We compare our results with the following baselines:
\begin{enumerate}
    \item Constant Velocity \cite{salzmann2020trajectron++}: Trajectories are predicted by setting acceleration to zero in the Verlet integration.
    \item Nearest Neighbour: We use a nearest neighbour search to find the closest absolute trajectory in the training set to the queried trajectory using an L2 metric. 
    \item STG-CNN \cite{mohamed2020social}: Uses a spatio-temporal graph convolutional neural network for trajectory prediction.  
    \item Transformer-TF  \cite{giuliari2021transformer}: Standard transformer implementation for trajectory prediction.  
\end{enumerate}
We chose not to include extensive comparative results from other social trajectory prediction algorithms. Our experiments found that the top-performing algorithms for pedestrian and AV benchmarks either exploit the problem's underlying structure or require non-trivial modifications, making them unsuitable for a domain transfer. A simple change in hyper-parameters very often did not provide optimal performance as can be seen from the short comparative result section. We use two popular metrics \cite{zhao2020noticing} for evaluating the performance of the proposed network: Average Displacement Error (ADE) and Final Displacement Error (FDE).
% \begin{enumerate}
%     \item Average Displacement Error (ADE): the L2 euclidean error distance over the entire predicted trajectory in absolute space averaged over all agents in that scene.
%     \item Final Displacement Error (FDE): the L2 euclidean error distance over the end point of the  predicted trajectory in absolute space averaged over all agents in that scene.
% \end{enumerate}
Results for the best of N trajectories are used where the network is queried N times, and the best ADE/FDE scores are recorded. We nominally use N = 5.
\subsection{Results and Discussions}

Figure \ref{fig:four_results} shows the qualitative results for the \textit{TrajAirNet}  framework. Four independent scenarios are showcased to highlight the various behaviours captured by \textit{TrajAirNet}. Figure \ref{fig:result_a} shows a scenario with two agents on the downwind leg of the pattern. The algorithm correctly predicts a longer turn to base for the trailing aircraft to improve spacing between agents. Figure \ref{fig:result_b} shows three aircraft with one aircraft trying to enter the pattern. While the algorithm correctly predicts the entry, it fails to predict the entering aircraft's roundabout turn to improve spacing. This highlights the diversity of maneuvers in the dataset and the difficulty in predicting them. The third scenario shows an aircraft entering downwind with another trailing aircraft already in the pattern. The network predicts that the trailing aircraft extends the crosswind leg and falls in line behind the aircraft entering the pattern. Lastly, the fourth scenario in Figure \ref{fig:result_d} shows three aircraft in the traffic pattern around the airport with the aircraft on downwind-to-base turning early to improve spacing. Table \ref{tab:results} shows the quantitative results for the \textit{TrajAirNet}  framework. As can be seen, \textit{TrajAirNet}  dominantly outperforms prediction baselines across all the sub-datasets.   

\section{Future Work }\label{sec:future}
% A major motivation of this work is to open up the world of general aviation to the wider robotics and automation community. With the recent advances in self-driving and social robotics, the aim of this publication is to encourage a more in-depth look into the problems faced by the aviation community and propose solutions for the same. 
One major drawback of the current work is the lack of generalizability across airports. More datasets covering multiple airports, including airports with control towers and multiple runways, are needed to augment this dataset. Similarly, the trajectory prediction network needs to generalize across different runway geometries. While the proposed trajectory prediction dataset and method provide a glimpse into the pilot's decision-making processes, pilots often use multiple data sources like radio communication and vision to supplement their situational awareness and predict other agents' behaviours. Future work involves collecting these concurrent multi-modal datasets and include the priors on terminal goals in the prediction framework. Transformers\cite{giuliari2021transformer} have shown promising results on various seq2seq tasks and need to be explored more in the trajectory prediction domain. 

\section{Conclusions}\label{sec:conclusion}

This work presents a novel dataset for socially-aware trajectory prediction in the aviation domain. Additionally, it also presents a baseline trajectory prediction algorithm for this static environment with a dynamic context. To the best of the authors' knowledge, this is the first publicly available dataset and method in the domain of general aviation. A major motivation of this work is to open up this domain to the wider robotics and automation community. With the recent advances in self-driving and social robotics, this publication aims to encourage a more in-depth look into the similar and unique problems faced by the autonomous aviation community and propose solutions in this relatively under-explored domain.

\addtolength{\textheight}{-3.5cm}   % This command serves to balance the column lengths
                                  % on the last page of the document manually. It shortens
                                  % the textheight of the last page by a suitable amount.
                                  % This command does not take effect until the next page
                                  % so it should come on the page before the last. Make
                                  % sure that you do not shorten the textheight too much.

%%%%%%%%%%%%%%%%%%%%%%%%%%%%%%%%%%%%%%%%%%%%%%%%%%%%%%%%%%%%%%%%%%%%%%%%%%%%%%%%

%%%%%%%%%%%%%%%%%%%%%%%%%%%%%%%%%%%%%%%%%%%%%%%%%%%%%%%%%%%%%%%%%%%%%%%%%%%%%%%%

%%%%%%%%%%%%%%%%%%%%%%%%%%%%%%%%%%%%%%%%%%%%%%%%%%%%%%%%%%%%%%%%%%%%%%%%%%%%%%%%

\section*{ACKNOWLEDGMENT}
The authors would like to thank the Pittsburgh-Butler Regional Airport for their support on this project especially to Airport Manager Richard E. ``Ike'' Kelly and Maintenance Manager Chuck Ritchey.
Authors would also like to thank Arti Anantharaman and Carlos Gonzalez for their support.

%%Added in title page
% This material is based upon work supported by the National Science Foundation Graduate Research Fellowship under Grant No. DGE1745016.

%%%%%%%%%%%%%%%%%%%%%%%%%%%%%%%%%%%%%%%%%%%%%%%%%%%%%%%%%%%%%%%%%%%%%%%%%%%%%%%%

\bibliography{ref}
\bibliographystyle{ieeetr}

\end{document}